\documentclass[sigconf]{acmart}

\usepackage{color}
\usepackage{booktabs}
\usepackage{float}
\usepackage{lineno}
\usepackage{multirow}
\usepackage{graphicx}
\usepackage[utf8]{inputenc}
\usepackage{algorithm}  
\usepackage{algpseudocode}  
\usepackage{amsmath} 
\usepackage{balance} 
\usepackage{appendix}
\usepackage{ulem}
\usepackage{enumitem}
\usepackage{graphicx}    % 插入图片必备
\usepackage{subcaption}  % 插入子图必备
\AtBeginDocument{%
  }

\copyrightyear{2025}
\acmYear{2025}
\setcopyright{acmlicensed}
\acmConference[CIKM '25] {Proceedings of the 34th ACM International Conference on Information and Knowledge Management}{ November 10--14, 2025}{Seoul, Republic of Korea.}
\acmBooktitle{Proceedings of the 34th ACM International Conference on Information and Knowledge Management (CIKM '25), November 10--14, 2025, Seoul, Republic of Korea}
\acmISBN{979-8-4007-2040-6/2025/11}
\acmDOI{10.1145/3746252.3760815}

\acmISBN{979-8-4007-2040-6/2025/11}

\begin{document}

\title{M3-Net: A Cost-Effective Graph-Free MLP-Based Model for Traffic Prediction}

\author{Guangyin Jin}
\authornote{Both authors contributed equally to this research.}
\affiliation{%
  \institution{National Innovative Institute of Defense Technology}\country{}}
\email{jinguangyin18@nudt.edu.cn}

\author{Sicong Lai}
\authornotemark[1]\affiliation{%
  \institution{The Hong Kong University of Science and Technology} \country{}}
\email{slai892@connect.hkust-gz.edu.cn}

\author{Xiaoshuai Hao}
\affiliation{%
  \institution{Beijing Academy of Artificial Intelligence} \country{}}
\email{xshao@baai.ac.cn}

\author{Mingtao Zhang}
\affiliation{%
  \institution{Peking University} \country{}}
\email{mingtaozhang@pku.edu.cn}

\author{Jinlei Zhang}
\authornote{Corresponding author}
\affiliation{%
  \institution{Beijing Jiaotong University}\country{}
}
\email{zhangjinlei@bjtu.edu.cn}

\renewcommand{\shortauthors}{Guangyin Jin, Sicong Lai, Xiaoshuai Hao, Mingtao Zhang, and Jinlei Zhang}

\begin{abstract}
Achieving accurate traffic prediction is a fundamental but crucial task in the development of current intelligent transportation systems. 
% Most of the mainstream methods that have made breakthroughs in traffic prediction rely on spatio-temporal graph neural networks, spatio-temporal attention mechanisms, etc. 
The main challenges of the existing deep learning approaches are that they either depend on a complete traffic network structure or require intricate model designs to capture complex spatio-temporal dependencies. 
These limitations pose significant challenges for the efficient deployment and operation of deep learning models on large-scale datasets.
To address these challenges, we propose a cost-effective graph-free Multilayer Perceptron (MLP) based model M3-Net for traffic prediction. 
% Our proposed model not only employs time series and spatio-temporal embeddings for efficient feature processing but also first introduces a novel MLP-Mixer architecture with a mixture of experts (MoE) mechanism. 
% The MLP-Mixer architecture jointly learns the spatial and channel dimensions of traffic flow data. In the spatial MLP module, adaptive grouped matrices are used to streamline spatial dimensions, thereby reducing the complexity of spatial learning. In the channel dimension, the MoE mechanism better captures multi-scale spatio-temporal dependencies. 
Extensive experiments conducted on multiple real datasets demonstrate the superiority of the proposed model in terms of prediction performance and lightweight deployment. 
Our code is available at {https://github.com/jinguangyin/M3\_NET}

\end{abstract}





\maketitle

\section{INTRODUCTION}

Accurate traffic forecasting serves as the analytical backbone of intelligent transportation systems.
% powering adaptive signal control, congestion pricing, fleet management, and mobility services. 
Recent advances in spatio-temporal (ST) models have significantly improved predictive accuracy by capturing how road segments interact and how these interactions evolve over time. 
However, modern urban networks are becoming increasingly large, dense, and noisy, widening the gap between laboratory benchmarks and real-world deployments. 
% \vspace{-10pt} % 根据需要调节 -5pt ~ -20pt
\begin{figure}[h]
    \setlength{\abovecaptionskip}{0pt}  % 图与标题之间的距离
    \setlength{\belowcaptionskip}{-5pt} % 标题与正文之间的距离
    \centering
    \vspace{-3mm}
    \includegraphics[width=0.9\linewidth]{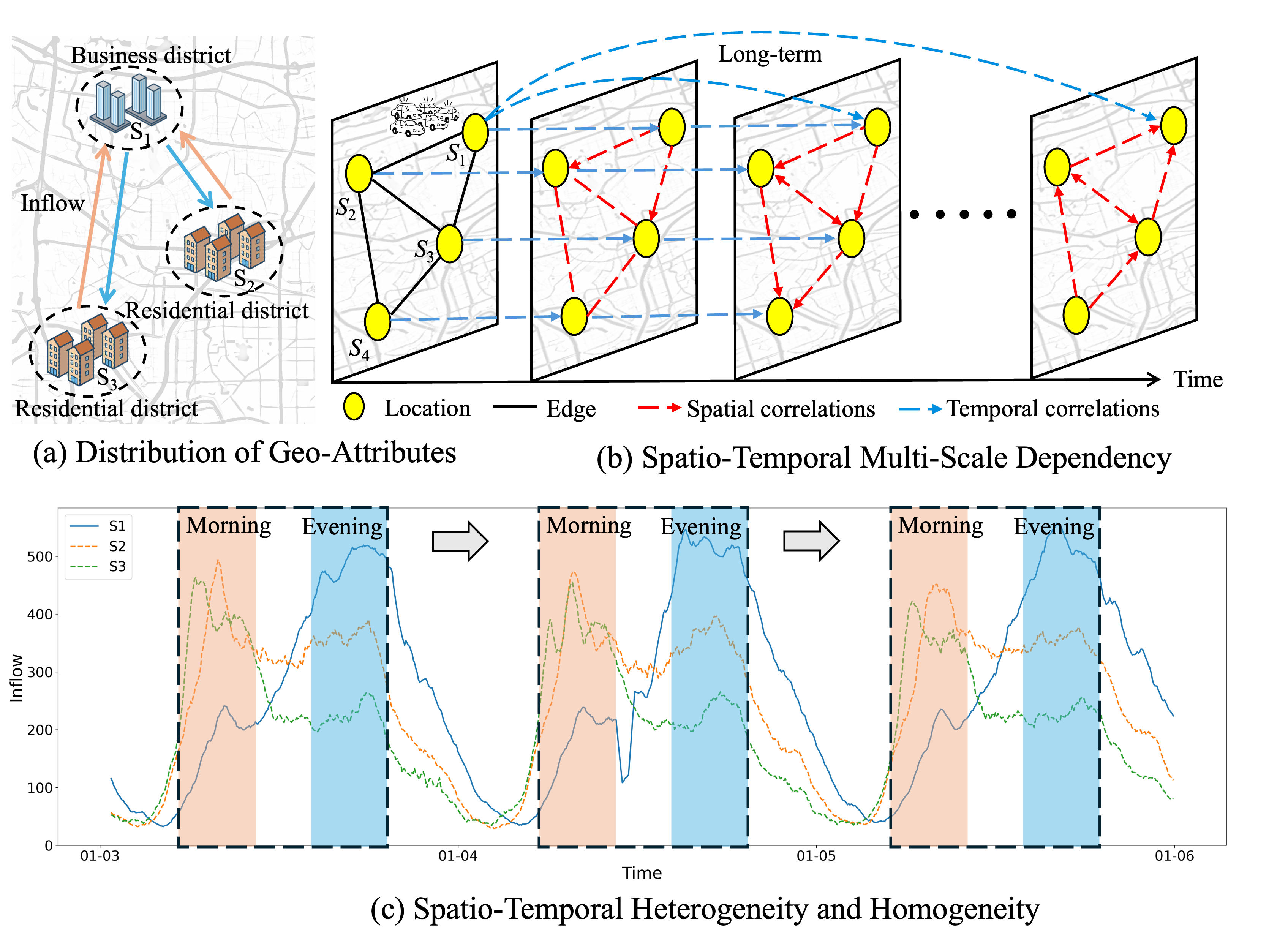} 
    \caption{Spatio-Temporal Characteristics of Traffic Flow.}
    \vspace{-2mm}
    \label{fig:figure1}
\end{figure}

This deployment gap highlights the structural complexity of traffic flow data. As illustrated in Figure~\ref{fig:figure1}, although S2 and S3 are geographically apart, both belong to residential districts and exhibit highly similar inflow patterns during morning and evening rush hours, demonstrating spatial homogeneity. In contrast, S1, located in a business district, shows significantly higher and more volatile inflow peaks in the morning, highlighting spatial heterogeneity across regions with different functional roles. The coexistence of homogeneity and heterogeneity poses challenges for modeling. 
% GNN-based methods can encode spatial correlations but depend on static graphs, which limits flexibility and increases computational cost. 
Moreover, Figure~\ref{fig:figure1}(b) illustrates the existence of complex multi-scale spatio-temporal dependencies in traffic flow. For example, a traffic disruption at node S1 may rapidly propagate to nearby nodes such as S2 and S3, and eventually extend to more distant areas like S4. This reflects both short-term local dependencies, characterized by immediate responses among adjacent regions, and long-term global dependencies, where the influence persists and spreads across space and time. Such patterns evolve dynamically across multiple spatial and temporal scales. 
% As a result, models based on fixed time windows or simplistic architectures often struggle to capture these hierarchical dependencies, limiting their effectiveness in real-world traffic forecasting. 
Therefore, it is essential to build models that can handle both spatial diversity and multi-scale spatio-temporal dependencies for accurate traffic forecasting.

Recently, although there has been significant growth of works for urban traffic prediction as well as analogous ST forecasting tasks~\cite{jin2023spatio,liu2023largest}, the aforementioned challenges are still not fully addressed. Some methods focus on modeling ST correlations by utilizing a single shared model across all locations~\cite{RNN,lstm,lai2018modeling}. Such approaches often overlook the inherent heterogeneity among different regions and fail to explicitly capture diverse spatial-temporal interaction patterns, making it difficult for the model to adapt to varying local characteristics. Another line of research builds graph-structured representations to enhance spatial modeling~\cite{stgcn,dcrnn,graphwavenet,jin2022automated,astgcn,jin2023stnpp,li2022automated}, typically propagating information along static or learned graphs. While effective in representing spatial relations, these models heavily rely on predefined topologies and incur significant computational overhead, which limits scalability in dynamic or large-scale networks. More recently, attention-based methods have been developed to learn long-range ST dependencies in a fully data-driven manner without explicit graph construction~\cite{attention,sttn,staeformer}. Although these methods improve modeling flexibility, they still struggle to distinguish between local heterogeneity and global homogeneity and often suffer from high memory and computational costs, hindering their deployment in real-world traffic systems.

To address these challenges, we propose \textbf{M3-Net}, a novel and efficient ST forecasting framework based on a pure MLP architecture. In contrast to traditional graph-structured or attention-based methods, M3-Net learns ST dynamics directly from raw traffic sequences without relying on explicit road network topology construction, thereby significantly reducing model complexity and improving deployment friendliness. Specifically, M3-Net introduces two key components. First, for spatial modeling, M3-Net designs a \textbf{Spatial MLP} module equipped with \textbf{adaptive grouping matrices}, which flexibly captures heterogeneous local patterns as well as homogeneous global structures, thus unifying the learning of complex spatial properties. Second, for temporal modeling, M3-Net introduces a \textbf{Channel MLP} enhanced with a \textbf{Mixture-of-Experts (MoE)} mechanism, enabling dynamic capacity allocation to better capture multi-scale ST dependencies present in traffic flows. Through this novel architecture, M3-Net achieves a favorable balance between modeling expressiveness, computational efficiency, and generalization, enabling stable and robust forecasting performance across urban networks of varying scales and structural complexities. 

\begin{figure*}[!ht]
    \centering
     \vspace{-3mm}
    \includegraphics[width=0.95\linewidth]{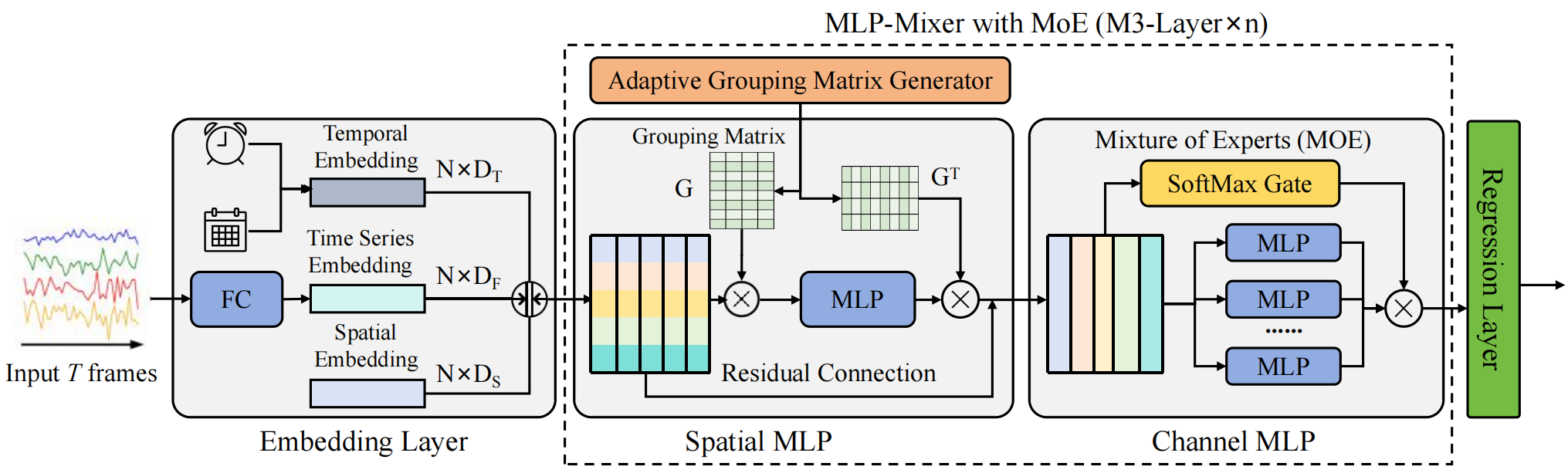}
    \caption{The framework of M3-Net.}
    \vspace{-3mm}
    \label{fig:pipeline}
\end{figure*}

% \section{PROBLEM DEFINITION}

\section{METHODOLOGY}
In this section, we introduce the proposed M3‑Net framework, as illustrated in Figure \ref{fig:pipeline}. Concretely, M3‑Net consists of four key building blocks as follows: 1) \textbf{Embedding Layer} fuses raw ST series with dedicated temporal (day‑time \& week‑time) and node embeddings to provide context‑rich token representations; 2) \textbf{M3 Layer} is composed of \textbf{Spatial MLP} and \textbf{Channel MLP}, where the Spatial MLP employs an adaptive grouping matrix to aggregate sensors into dynamic clusters and applies token mixing to capture inter‑cluster interactions with residual refinement, the Channel MLP is with a MoE mechanism, enabling the model to learn heterogeneous feature dependencies while remaining lightweight; 3) \textbf{Regression Layer} performs linear projection to produce accurate multi‑horizon traffic predictions.

\subsection{Embedding Layer}
To capture the complex spatio-temporal dependencies in traffic sequences, we embed the raw input data using a fully connected layer. Given the historical traffic sequence $X_{t-L+1:t} \in \mathbb{R}^{L \times N \times C}$, where $L$ is the sequence length, $N$ is the number of nodes, and $C$ is the number of input features. We obtain the dynamic feature embedding $E_F \in \mathbb{R}^{N \times D_F}$ via $E_F = \mathrm{FC}(X_{t-L+1:t})$, where $D_F$ is the projected feature dimension. To model node-specific static characteristics, we introduce a learnable spatial embedding $E_S \in \mathbb{R}^{N \times D_S}$. Meanwhile, to incorporate temporal periodicity, we construct two temporal embedding dictionaries: a time-of-day embedding matrix $E_d \in \mathbb{R}^{T_d \times D_d}$ and a day-of-week embedding matrix $E_w \in \mathbb{R}^{T_w \times D_w}$, where $T_d$ and $T_w$ denote the number of intervals in a day and week, respectively. The corresponding temporal embeddings are retrieved via index lookups: $E_T^{(d)} = E_d[\text{idx}_t^{(d)}] \in \mathbb{R}^{N \times D_d}$ and $E_T^{(w)} = E_w[\text{idx}_t^{(w)}] \in \mathbb{R}^{N \times D_w}$. Finally, we concatenate all embeddings along the feature dimension to form the complete spatiotemporal representation $H \in \mathbb{R}^{N \times D_H}$, where $H = E_F \parallel E_S \parallel E_T^{(d)} \parallel E_T^{(w)}$ and $D_H = D_F + D_S + D_d + D_w$. 
% This fused embedding tensor $H$ is subsequently fed into the MLP-Mixer with Mixture-of-Experts for spatio-temporal modeling.

\subsection{MLP-Mixer with MoE}
\textbf{Spatial MLP.} To model spatial dependencies among nodes, we introduce a Spatial MLP module equipped with an adaptive grouping mechanism. Given the input tensor $H \in \mathbb{R}^{N \times D}$, we generate an adaptive grouping matrix $G \in \mathbb{R}^{N \times g}$ to partition the $N$ nodes into $g$ groups. This matrix enables aggregation of node-level features into group-level representations via $H_g = G^\top H \in \mathbb{R}^{g \times D}$, where $G^\top$ denotes the transpose of $G$. The grouped features are then passed through a shared MLP to capture intra-group patterns, yielding $\hat{H}_g = \mathrm{MLP}(H_g) \in \mathbb{R}^{g \times D}$. These transformed features are subsequently mapped back to the original node space by multiplying with $G$, resulting in $\hat{H} = G \hat{H}_g \in \mathbb{R}^{N \times D}$. A residual connection is then added to preserve the original representation, giving the spatially enhanced output $H_s = H + \hat{H}$. 
% This design allows the model to learn localized spatial correlations efficiently while maintaining the flexibility to adaptively group nodes.

\textbf{Channel MLP.} To further capture nonlinear interactions along the feature dimension, we introduce a Channel MLP module with a Mixture-of-Experts (MoE) mechanism. The spatial-enhanced output $H_s \in \mathbb{R}^{N \times D}$ is first passed through a softmax-based gating network to compute expert weights $\alpha = \mathrm{SoftmaxGate}(H_s) \in \mathbb{R}^{N \times K}$, where $K$ denotes the number of experts. Each expert $\mathrm{MLP}_k$ independently processes the same input to produce $O_k = \mathrm{MLP}_k(H_s) \in \mathbb{R}^{N \times D}$ for $k = 1, \dots, K$. These outputs are then weighted and combined using the gating scores, resulting in the final output representation $H_c = \sum_{k=1}^{K} \alpha_k \odot O_k$, where $\odot$ denotes element-wise multiplication. 
% This expert-based architecture dynamically allocates modeling capacity across different feature subspaces, enhancing the model’s expressiveness and its ability to adapt to diverse input patterns.

\subsection{Regression Layer}
After embedding layer and M3 layer, we obtain a representation tensor $H_c \in \mathbb{R}^{N \times D}$ enriched with deep spatio-temporal dependencies. The prediction head is responsible for transforming this latent representation into the final forecasting output. 
% Specifically, it first aggregates the temporal information across the sequence length $L$ for each spatial node, yielding a compressed representation $Z \in \mathbb{R}^{N \times D}$, typically via temporal average pooling or attention-based summarization. This aggregated tensor is then fed into a fully connected regression layer $FC_{\text{regression}}$ that maps each node’s encoded features to its corresponding future value predictions over a horizon of length $F$. 
The final forecasting output is thus computed as $\hat{Y}_{t:t+F} = FC_{\text{regression}}(H_c)$.

% where $Z_t$ represents the temporally pooled representation for all nodes and $\hat{Y}_{t:t+F}\in \mathbb{R}^{N \times F}$ is the predicted value sequence for each node over the next $F$ time steps. 

\section{EXPERIMENTS}

\subsection{Experimental Setup}
We evaluate our model on four normal datasets and three large-scale datasets. The normal datasets include PEMS03, PEMS04, PEMS07 and PEMS08, Which are collected from California highways~\cite{song2020spatial}.
% The large-scale datasets include SD, GLA and GBA, which are collected from larger interstate highways in the United States~\cite{liu2023largest}. 
% The detailed statistics of the two datasets are shown in Table~\ref{tab:data}.
% Each dataset is chronologically split with 60\% for training, 20\% for validation and 20\% for testing.
In the experiments, we use the traffic flow of the last 12 time steps to predict the traffic flow of the next 12 time steps, and record the prediction performance of the 3rd, 6th, 12th steps and the average. Our model is implemented by Pytorch 1.7 with vGPU-32GB. 
We set the number of spatial groups as 10 in spatial MLP, the number of experts in channel MLP as 4 and the number of M3-Layers as 3 by default. The dimension of hidden representations in our model is set as 128 and the dimension of embedding layer is set as 32. 
We set the Adam optimizer with an initial learning rate of 0.002, where the learning rate follows a step-wise decay strategy, and the batch size is set as 64. 
% During the training phase, we employ the early stopping strategy with tolerance 30 for 200 epochs.
\begin{table}[!ht]
\centering
  \vspace{-3mm}
% \small
\tabcolsep=0.7mm
  \caption{The description of the datasets in the experiments.}
  \scalebox{1.0}{
  \begin{tabular}{c|ccccc}
    \hline
     Dataset & Nodes & Edges & Time Range & Frames \\
    \hline \hline
    \multirow{4}{*}{}
     PEMS03 & 358 & 546  & 09/01/2018 -- 11/30/2018 & 26,208  \\
     PEMS04 & 307 & 338 & 01/01/2018 -- 02/28/2018 & 16,992  \\
     PEMS07 & 883 & 865 & 05/01/2017 -- 08/06/2017 & 28,224 \\
     PEMS08 & 170 & 276 & 07/01/2016 -- 08/31/2016 & 17,856  \\
    \hline
  \end{tabular}}
  \label{tab:data}
  \vspace{-3mm}
\end{table}

\begin{table*}
\renewcommand\arraystretch{1}
    \centering
\vspace{-3mm}
    \setlength{\abovecaptionskip}{0.cm}
    \setlength{\belowcaptionskip}{-0.0cm}
    \caption{Performance comparisons on normal datasets. We bold the best-performing results and underline the suboptimal results. }
    \label{tab:main}
\scalebox{0.7}{
    \begin{tabular}{cc|cccc|cccc|cccc|cccc}
    \toprule
    \midrule
    \multicolumn{2}{c|}{\textbf{Dataset}} & \multicolumn{4}{c|}{\textbf{PEMS04}} & \multicolumn{4}{c|}{\textbf{PEMS07}} & \multicolumn{4}{c|}{\textbf{PEMS08}}& \multicolumn{4}{c|}{\textbf{PEMS03}}\tabularnewline
    \midrule
    \midrule
    \textbf{Method} & \textbf{Metric} & \textbf{@3} & \textbf{@6} & \textbf{@12} & \textbf{Avg.} & \textbf{@3} & \textbf{@6} & \textbf{@12} & \textbf{Avg.} & \textbf{@3} & \textbf{@6} & \textbf{@12} & \textbf{Avg.} & \textbf{@3} & \textbf{@6} & \textbf{@12} & \textbf{Avg.}\tabularnewline
    \midrule
% \multirow{3}{*}{VAR} & MAE & 21.94 & 23.72 & 26.76 & 23.51 & 32.02 & 35.18 & 38.37 & 37.06 & 19.52 & 22.25 & 26.17 & 22.07 & 18.69 & 20.53 & 22.92 & 20.49\tabularnewline
%  & RMSE & 34.30 & 36.58 & 40.28 & 36.39 & 48.83 & 52.91 & 56.82 & 55.73 & 29.73 & 33.30 & 38.97 & 31.02 & 31.09 & 34.16 & 38.31 & 34.08 \tabularnewline
%  & MAPE & 16.42\% & 18.02\% & 20.94\% & 17.85\% & 18.30\% & 20.54\% & 22.04\% & 19.93\% & 12.54\% & 14.23\% & 17.32\% & 14.04\% & 21.38\% & 22.27\% & 24.37\% & 22.17\% \tabularnewline
% \midrule
% \multirow{3}{*}{LSTM} & MAE & 21.37 & 23.72 & 26.76 & 23.81 & 20.42 & 23.18 & 28.73 & 23.54 & 17.38 & 21.22 & 30.69 & 21.31 & 16.18 & 17.05 & 18.87 & 17.02\tabularnewline
%  & RMSE & 33.31 & 36.58 & 40.28 & 36.62 & 33.21 & 37.54 & 45.63 & 38.20 & 26.27 & 31.97\% & 43.96 & 32.10 & 26.72 & 28.91 & 30.76 & 28.95 \tabularnewline
%  & MAPE & 15.21\% & 18.02\% & 20.94\% & 18.12\% & 8.79\% & 9.80\% & 12.23\% & 9.96\% & 12.63\% & 17.32\% & 25.72\% & 17.47\% & 16.39\% & 17.78\% & 18.69\% & 17.84\% \tabularnewline
% \midrule
\multirow{3}{*}{DCRNN} & MAE & 18.53 & 19.65 & 21.67 & 19.71 & 19.45 & 21.18 & 24.14 & 21.20 & 14.16 & 15.24 & 17.70 & 15.26 & 14.41 & 15.52 & 16.77 & 15.54  \tabularnewline
 & RMSE & 29.61 & 31.37 & 34.19 & 31.43 & 31.39 & 34.42 & 38.84 & 34.43 & 22.20 & 24.26 & 27.14 & 24.28 & 25.10 & 27.15 & 29.63 & 27.18\tabularnewline
 & MAPE & 12.71\% & 13.45\% & 15.03\% & 13.54\% & 8.29\% & 9.01\% & 10.42\% & 9.06\% & 9.31\%\% & 9.90\% & 11.13\% & 9.96\% & 14.43 & 15.58\% & 17.24\% & 15.62\% \tabularnewline
\midrule
\multirow{3}{*}{STGCN} & MAE & 18.74 & 19.64 & 21.12 & 19.63 & 20.33 & 21.66 & 24.16 & 21.71 & 14.95 & 15.92 & 17.65 & 15.98 & 14.85 & 15.79 & 17.79 & 15.83\tabularnewline
 & RMSE & 29.84 & 31.34 & 33.53 & 31.32 & 32.73 & 35.35 & 39.48 & 35.41 & 23.48 & 25.36 & 28.03 & 25.37 & 25.67 & 27.48 & 30.16 & 27.51 \tabularnewline
 & MAPE & 14.42\% & 13.27\% & 14.22\% & 13.32\% & 8.68\% & 9.16\% & 10.26\% & 9.25\% & 9.87\% & 10.42\% & 11.34\% & 10.43\% & 14.62\% & 16.11\% & 17.67\% & 16.13\% \tabularnewline
\midrule
\multirow{3}{*}{GWNet} & MAE & 18.00 & 18.96 & 20.53 & 18.97 & 18.69 & 20.26 & 22.79 & 20.25 & 13.72 & 14.67 & 16.15 & 14.67 & \textbf{13.51} & \textbf{14.68} & \textbf{15.71} & \textbf{14.63} \tabularnewline
 & RMSE & 28.83 & 30.33 & 32.54 & 30.32 & 30.69 & 33.37 & 37.11 & 33.32 & 21.71 & \uline{23.50} & 25.95 & 23.49 & \textbf{23.41} & \textbf{25.32} & \textbf{27.94} & \textbf{25.24} \tabularnewline
 & MAPE & 13.64\% & 14.23\% & 15.41\% & 14.26\% & 8.02\% & 8.56\% & 9.73\% & 8.63\% & 8.80\% & 9.49\% & 10.74\% & 9.52\% & 14.32\% & 15.63\% & 17.03\% & 15.52\% \tabularnewline
\midrule
\multirow{3}{*}{AGCRN} & MAE & 18.52 & 19.45 & 20.64 & 19.36 & 19.31 & 20.70 & 22.74 & 20.64 & 14.51 & 15.66 & 17.49 & 15.65 & 14.13 & 15.23 & 16.58 & 15.24 \tabularnewline 
& RMSE & 29.79 & 31.45 & 33.31 & 31.28 & 31.68 & 34.52 & 37.94 & 34.39 & 22.87 & 25.00 & 27.93 & 24.99 & 24.32 & 26.47 & 28.52 & 26.65 \tabularnewline
& MAPE & 12.31\% & 12.82\% & 13.74\% & 12.81\% & 8.18\% & 8.66\% & 9.71\% & 8.74\% & 9.34\% & 10.34\% & 11.72\% & 10.17\% & 14.75\% & 15.92\% & 17.11\% & 15.89\% \tabularnewline
\midrule
\multirow{3}{*}{StemGNN} & MAE & 19.48 & 21.40 & 24.90 & 21.61 & 19.74 & 22.07 & 26.20 & 22.23 & 14.49 & 15.84 & 18.10 & 15.91 & 15.86 & 16.91 & 18.10 & 16.95 \tabularnewline
 & RMSE & 30.74 & 33.46 & 38.29 & 33.80 & 32.32 & 36.16 & 42.32 & 36.46 & 23.02 & 25.38 & 28.77 & 25.44 & 26.25 & 28.54 & 30.13 & 28.52 \tabularnewline
 & MAPE & 13.84\% & 15.85\% & 19.50\% & 16.10\% & 8.27\% & 9.20\% & 11.00\% & 9.20\% & 9.73\% & 10.78\% & 12.50 & 10.90\% & 17.84\% & 19.51\% & 21.02\% & 19.61\% \tabularnewline
\midrule
\multirow{3}{*}{GMAN} & MAE & 18.27 & 18.81 & 20.01 & 18.83 & 19.25 & 20.33 & 22.25 & 20.43 & 13.80 & 14.62 & 15.72 & 14.81 & 14.03 & 15.17 & 16.20 & 15.14 \tabularnewline
 & RMSE & 29.35 & 30.85 & \textbf{31.32} & 30.93 & 31.20 & 33.30 & 36.40 & 33.30 & 22.88 & 24.12 & 26.47 & 24.19 & 24.65 & 26.23 & 28.95 & 26.15 \tabularnewline
 & MAPE & 12.66\% & 13.25\% & 13.40\% & 13.21\% & 8.21\% & 8.63\% & 9.48\% & 8.69\% & 9.41\% & 9.57\% & 10.56\% & 9.69\% & 14.31\% & 15.38\% & 16.98\% & 15.41\% \tabularnewline
\midrule
\multirow{3}{*}{MTGNN} & MAE & 18.65 & 19.48 & 20.96 & 19.50 & 19.23 & 20.83 & 23.60 & 20.94 & 14.30 & 15.25 & 16.80 & 15.31 & 13.74 & 14.97 & 16.01 & 14.92\tabularnewline
& RMSE & 30.13 & 32.02 & 34.66 & 32.00 & 31.15 & 33.93 & 38.10 & 34.03 & 22.55 & 24.41 & 26.96 & 24.42 & 23.60 & 25.82 & 28.26 & 25.53 \tabularnewline
& MAPE & 13.32\% & 14.08\% & 14.96\% & 14.04\% & 8.55\% & 9.30\% & 10.10\% & 9.10\% & 10.56\% & 10.54\% & 10.90\% & 10.70\% & 13.87\% & \textbf{14.49\%} & 16.40\% & \textbf{14.55\%} \tabularnewline
\midrule
\multirow{3}{*}{DGCRN} & MAE & 18.85 & 20.04 & 22.32 & 20.29 & 19.03 & 20.41 & 22.58 & 20.44 & 13.79 & 14.81 & 16.39 & 14.85 & 13.76 & 14.93 & 16.04 & 14.95\tabularnewline
& RMSE & 29.95 & 32.07 & 36.28 & 32.55 & 30.74 & 33.27 & 36.74 & 33.25 & 21.91 & 23.83 & 26.34 & 23.84 & 23.71 & 25.78 & 28.31 & 25.62 \tabularnewline
& MAPE & 12.92\% & 13.50\% & 14.61\% & 13.60\% & 8.16\% & 8.69\% & 9.63\% & 8.73\% & 9.13\% & 9.74\% & 11.02\% & 9.84\% & 14.17\% & 15.23\% & 16.98\% & 15.35\% \tabularnewline
\midrule
\multirow{3}{*}{STNorm} & MAE & 18.28 & 18.92 & 20.20 & 18.96 & 19.15 & 20.63 & 22.60 & 20.52 & 14.44 & 15.53 & 17.20 & 15.54 & 14.21 & 15.32 & 16.28 & 15.35 \tabularnewline
 & RMSE & 29.70 & 31.12 & 32.91 & 30.98 & 31.70 & 35.10 & 38.65 & 34.85 & 22.68 & 25.07 & 27.86 & 25.01 & 23.62 & 25.88 & 28.27 & 25.93 \tabularnewline
 & MAPE & 12.28\% & 12.71\% & 13.43 & 12.69\% & 8.26\% & 8.84\% & 9.60\% & 8.77\% & 9.32\% & 9.98\% & 11.30\% & 10.03\% & \textbf{13.85\%} & 14.53\% & \textbf{16.36\%} & 14.56\% \tabularnewline
\midrule
\multirow{3}{*}{STID} & MAE & \uline{17.62} & \uline{18.43} & \uline{19.82} & \uline{18.44} & \uline{18.31} & \uline{19.59} & \uline{21.52} & \uline{19.54} & \uline{13.28} & \uline{14.21} & \uline{15.58} & \uline{14.20} & 14.17 & 15.29 & 16.27 & 15.31 \tabularnewline
 & RMSE & \textbf{28.57} & \textbf{29.93} & \uline{31.95} & \textbf{29.92} & \uline{30.39} & \uline{32.90} & \uline{36.29} & \uline{32.85} & \uline{21.66} & 23.57 & \uline{25.89} & \uline{23.49} & 24.81 & 27.36 & 29.79 & 27.40 \tabularnewline
 & MAPE & \textbf{12.00\%} & \uline{12.52\%} & \uline{13.63\%} & \uline{12.58\%} & \uline{7.72\%} & \uline{8.30\%} & \uline{9.15\%} & \uline{8.25\%} & \uline{8.62\%} & \uline{9.24\%} & \uline{10.33\%} & \uline{9.28\%} & 14.72\% & 16.43\% & 17.65\% & 16.40\% \tabularnewline
\midrule
\multirow{3}{*}{M3-Net} & MAE & \textbf{17.58} & \textbf{18.34} & \textbf{19.42} & \textbf{18.30} & \textbf{18.18} & \textbf{19.51} & \textbf{21.28} & \textbf{19.42} & \textbf{13.05} & \textbf{13.95} & \textbf{15.01} & \textbf{13.89} & \uline{13.69} & \uline{14.92} & \uline{15.86} & \uline{14.88} \tabularnewline
 & RMSE & \uline{28.82} & \uline{30.13} & \textbf{31.82} & \uline{30.15} & \textbf{30.21} & \textbf{32.78} & \textbf{36.03} & \textbf{32.85} & \textbf{21.48} & \textbf{23.47} & \textbf{25.54} & \textbf{23.33} & \uline{23.58} & \uline{25.77} & \uline{28.19} & \uline{25.49} \tabularnewline
 & MAPE & \uline{12.10\%} & \textbf{12.44\%} & \textbf{13.41\%} & \textbf{12.53\%} & \textbf{7.66\%} & \textbf{8.20\%} & \textbf{9.07\%} & \textbf{8.21\%} & \textbf{8.52\%} & \textbf{9.14\%} & \textbf{10.02\%} & \textbf{9.17\%} & \uline{13.98\%} & \uline{14.99\%} & \uline{16.82\%} & \uline{15.02\%} \tabularnewline 
 \midrule
\bottomrule
\end{tabular}}
\label{tab:normal}
 \vspace{-3mm}
\end{table*}

\subsection{Performance Evaluation}
We compare our proposed model with fourteen state-of-art baselines as follows: DCRNN~\cite{dcrnn}, STGCN~\cite{stgcn}, GWNet~\cite{gwn}, AGCRN~\cite{agcrn}, StemGNN~\cite{stemgnn}, GMAN~\cite{gman}, MTGNN~\cite{mtgnn}, DGCRN~\cite{dgcrn}, STNorm~\cite{stnorm} and STID~\cite{stid}.
% ASTGCN~\cite{astgcn}, STGODE~\cite{stgode} and DSTAGNN~\cite{dstagnn}. 
The evaluation metrics are mean absolute errors (MAE), root mean squared errors (RMSE) and mean absolute percentage errors (MAPE) averaged over five times for one hour ahead prediction. 
The comparison results of our proposed model with other baselines on normal datasets are shown in Table~\ref{tab:normal}. We can observe that our model achieves the best results on most metrics across multiple datasets and also achieves the second-best results on the remaining metrics. Baselines such as GWNet, MTGNN, AGCRN, StemGNN and DGCRN are commonly used state-of-the-art models based on adaptive graphs in the field of ST prediction. The overall performance of M3-Net surpasses these models, demonstrating that the combination of the spatio-temporal embeddings and a finely designed MLP architecture can outperform some more complexly designed models. 
% Furthermore, M3-Net also outperforms the recently proposed graph-free models STNorm and STID. Notably, STID is also an excellent model based on MLP, and the reasons why M3-Net surpasses STID can be attributed to two main aspects: first, M3-Net learns spatial homogeneity information better through the design of spatial MLP; second, the MoE mechanism partially compensates for the limitations of simple MLP models in learning spatio-temporal multi-scale information. 
% We also conducted comparative experiments on three large-scale datasets, as shown in Table~\ref{tab:largest}. From the experimental results, we can observe that our proposed M3-Net outperforms other optimal baseline models even in large-scale scenarios. Additionally, considering M3-Net's high efficiency and linear complexity, these factors provide a solid foundation for deploying the proposed model on large-scale real-world datasets.

\begin{table}[!htb]
\caption{Ablation experiments.}
 \vspace{-3mm}
\scalebox{1.0}{
	\begin{tabular}{clccc}
		\hline
		Dataset                 & Model\&Variants   & MAE & RMSE           & MAPE   \\ \hline
		\multirow{4}{*}{PEMS04} & M3-Net        & \textbf{18.30} & \textbf{30.15}  & \textbf{12.53\%}          \\
		& w/o MOE   & 18.39 & 30.36 & 12.67\% \\
		& w/o Spatial MLP & 18.42 & 30.31 & 12.63\% \\
		& w/o Grouping Matrix  & 18.83 & 30.95 & 12.76\%  \\ \hline
		\multirow{4}{*}{PEMS08} & M3-Net  & \textbf{13.89} & \textbf{23.33}  & \textbf{9.17\%}         \\
		& w/o MOE   & 14.04 & 23.42 & 9.25\%  \\
		& w/o Spatial MLP & 14.10 & 23.45 & 9.23\%  \\
		& w/o Grouping Matrix  & 13.95 & 23.51 & 9.21\%  \\ \hline
	\end{tabular}}
	\label{tab:ablation}
 	\vspace{-3mm}
\end{table}

\subsection{Ablation Study}
We conduct ablation study on PEMS04 and PEMS08 to evaluate the effectiveness of each crucial module in our model. As shown in Table \ref{tab:ablation}, we compared M3-Net with following ablation variants: 1) \emph{w/o MOE}, which removes the mixture of experts mechanism from channel MLP module 2) \emph{w/o Spatial MLP}, which removes the spatial MLP module from our model.   
3) \emph{w/o Group Matrix}, which removes the adaptive grouping matrix from spatial MLP module. 

From Table~\ref{tab:ablation}, we can find that our complete model M3-Net outperforms all the ablation variants. 
% Compared with the results of \emph{w/o MOE}, M3-Net improves 0.5\%, 0.7\%, 1.1\% in terms of MAE, RMSE and MAPE on PEMS04, and it improves 1.1\%, 0.4\%, 0.8\% in terms of the three metrics on PEMS08. 
The comparison of \emph{w/o MOE} can demonstrate that capturing multi-scale spatio-temporal dependencies of traffic flow in road networks can be beneficial for predictive performance. 
% Compared with the results of \emph{w/o Spatial MLP}, M3-Net improves 0.7\%, 0.7\%, 0.8\% in terms of MAE, RMSE and MAPE on PEMS04, and it improves 1.5\%, 0.5\%, 0.7\% in terms of the three metrics on PEMS08. 
The variant of \emph{w/o Spatial MLP} can illustrate that the spatial MLP can capture the spatial correlations in historical traffic sequences for more accurate prediction. 
To further investigate the effectiveness of adaptive group matrix, we compare M3-Net with \emph{w/o Grouping Matrix}. 
% Our model improves 2.9\%, 2.7\%, 1.8\% in terms of MAE, RMSE and MAPE on PEMS04, and it improves 0.4\%, 0.8\%, 0.4\% in terms of the three metrics on PEMS08. 
This comparison indicates that capturing spatial homogeneity is crucial for prediction performance. 
% If the adaptive grouping matrix learning module is removed, it may not only prevent the learning of spatial homogeneity features but also lead to performance degradation due to overfitting.

% \subsection{Case Study}
\subsection{Spatial Homogeneity Visualization}
\begin{figure}[!t]
    \centering
    \vspace{-1mm}
    \includegraphics[width=1\linewidth]{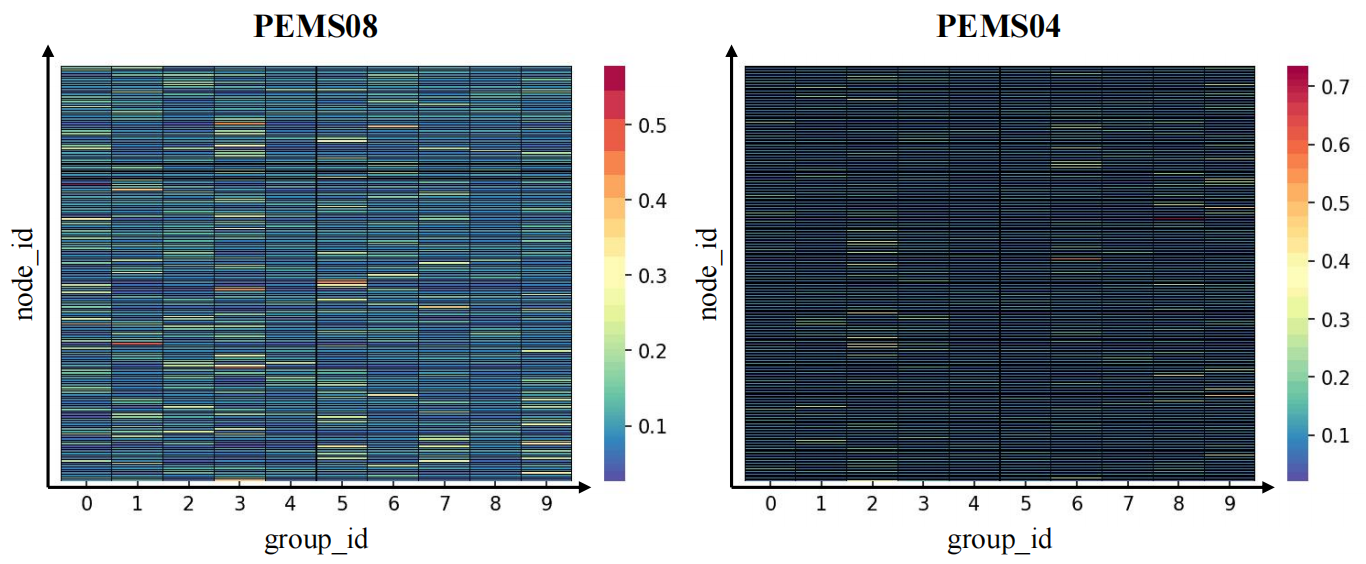}
    \caption{Adaptive grouping matrices visualization on PEMS04 and PEMS08.}
    \vspace{-3mm}
    \label{fig:matrix}
\end{figure}

To further investigate how our proposed model characterizes the spatial homogeneity by the adaptive grouping matrix, we visualize the learned matrices on PEMS04 and PEMS08 by heatmaps. As shown in Fig.~\ref{fig:matrix}, 
the vertical axis represents the index of spatial nodes, and the horizontal axis represents the index of groups. In the heatmaps, colors closer to warm tones in each cell indicate a higher compatibility between a node and a group, whereas colors closer to cool tones indicate lower compatibility. It is easy to observe that the compatibility of some nodes with certain groups significantly exceeds the average value. Therefore, the adaptive grouping matrices effectively learn spatial homogeneity features. 
% Additionally, this probabilistic soft constraint allows some nodes to have strong associations with multiple different groups, thereby preserving the diversity of homogeneity features.

\subsection{Cost-Effectiveness Analysis}
\begin{figure}[!t]
    \centering
    \vspace{-1mm}
    \includegraphics[width=1\linewidth]{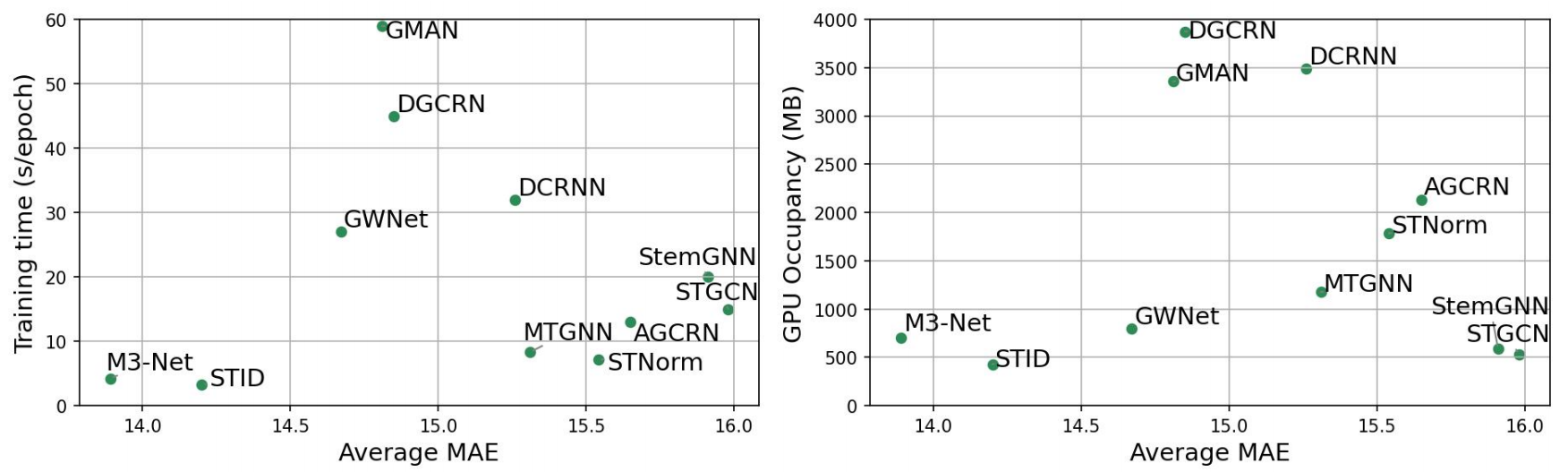}
    \caption{Visualization of cost-effectiveness of deployment on PEMS08 dataset.}
    \vspace{-3mm}
    \label{fig:cost}
\end{figure}

Time consumption and GPU memory usage are two intuitive indicators reflecting the time and space complexity of different models. Based on these two indicators, we conducted a comparative analysis of the cost-effectiveness of the proposed model and other baseline models on PEMS08 dataset. As shown in Fig.~\ref{fig:cost}, the horizontal axis represents the average MAE metric values of model predictions, while the vertical axis represents the training time per epoch and the GPU memory usage. The closer the scatter points are to the origin, the higher the cost-effectiveness of the model, indicating that it balances prediction accuracy with efficiency and deployment cost. Observing the experimental results, it is evident that the cost-effectiveness of M3-Net surpasses other baseline models in terms of training time and memory usage, especially outperforming some previous state-of-the-art models based on spatio-temporal graphs. 
% This superiority is attributed to the lightweight MLP backbone of M3-Net, which captures both heterogeneous and homogenous spatial node features through spatial embeddings and adaptive grouped matrices in spatial learning, without the need for introducing graph structures that significantly increase computational overhead. The findings of the cost-effectiveness analysis suggest that the M3-Net model has significant potential for practical deployment and application on large-scale datasets.

\section{CONCLUSION}
In this paper, we propose M3-Net, a lightweight and effective graph-free Multilayer Perceptron model to addresses the critical need for accurate traffic prediction in intelligent transportation systems. The innovative MLP-Mixer architecture, enhanced with a mixture of experts mechanism, allows for the joint learning of spatial and channel dimensions in traffic flow data. The incorporation of adaptive grouped matrices within the spatial MLP module effectively reduces the complexity of spatial learning, facilitating a more streamlined and efficient model.
The extensive experiments conducted on various real-world datasets validate the effectiveness of M3-Net, demonstrating its superior performance in traffic prediction while maintaining a lightweight and deployable structure. 
Future work will explore further enhancements to the M3-Net architecture and its applicability to other domains within transportation and urban planning. 
% paving the way for more robust and scalable solutions in traffic prediction and management.

% \begin{acks}
% \end{acks}

\section{GenAI Usage Disclosure}
None of the text, data, and code associated with this paper was produced by generative AI. We only used generative AI for language polishing.

\bibliographystyle{ACM-Reference-Format}
\bibliography{sample-base}

\appendix

\end{document}